\documentclass[10pt,twocolumn,letterpaper]{article}

\usepackage{cvpr}
\usepackage{times}
\usepackage{epsfig}
\usepackage{graphicx}
\usepackage{amsmath}
\usepackage{amssymb}
\usepackage{multirow}
\usepackage{pifont}
\usepackage{mathrsfs}
\usepackage{bbding}
\usepackage{color}

\usepackage{xcolor}

\definecolor{mycyan}{RGB}{0,255,255}
\definecolor{mymagenta}{RGB}{255,0,255}

\newcommand{\cfbox}[2]{%
\colorlet{currentcolor}{.}%
{\color{#1}%
	\fbox{\color{currentcolor}#2}}%
}

\usepackage[breaklinks=true,bookmarks=false]{hyperref}

\cvprfinalcopy % *** Uncomment this line for the final submission

\def\cvprPaperID{6780} % *** Enter the CVPR Paper ID here

\ifcvprfinal\fi
\begin{document}

%%%%%%%%% TITLE

\title{Cross-Domain Document Object Detection: Benchmark Suite and Method}

\author{Kai Li$^{1}$, Curtis Wigington$^{2}$, Chris Tensmeyer$^{2}$, Handong Zhao$^{2}$, \\  Nikolaos Barmpalios$^{3}$, Vlad I. Morariu$^{2}$, Varun Manjunatha$^{2}$, Tong Sun$^{2}$, Yun Fu$^{1}$ \\
$^1$Northeastern University, $^2$Adobe Research, $^3$Adobe Document Cloud  \\
% $^4$Khoury College of Computer Science, Northeastern University, Boston, USA \\
{\tt\small \{kaili,yunfu\}@ece.neu.edu} \\ {\tt\small\{wigingto,tensmeye,hazhao,barmpali,morariu,vmanjuna,tsun\}@adobe.com}
}

% \author{First Author\\
% 	Institution1\\
% 	Institution1 address\\
% 	{\tt\small firstauthor@i1.org}
% 	% For a paper whose authors are all at the same institution,
% 	% omit the following lines up until the closing ``}''.
% 	% Additional authors and addresses can be added with ``\and'',
% 	% just like the second author.
% 	% To save space, use either the email address or home page, not both
% 	\and
% 	Second Author\\
% 	Institution2\\
% 	First line of institution2 address\\
% 	{\tt\small secondauthor@i2.org}
% }

\maketitle
\thispagestyle{empty}

%%%%%%%%% ABSTRACT
\begin{abstract}		
	Decomposing images of document pages into high-level semantic regions (e.g., figures, tables, paragraphs), document object detection (DOD) is fundamental for downstream tasks like intelligent document editing and understanding. DOD remains a challenging problem as document objects vary significantly in layout, size, aspect ratio, texture, etc. An additional challenge arises in practice because large labeled training datasets are only available for domains that differ from the target domain. We investigate cross-domain DOD, where the goal is to learn a detector for the target domain using labeled data from the source domain and only unlabeled data from the target domain. Documents from the two domains may vary significantly in layout, language, and genre. We establish a benchmark suite consisting of different types of PDF document datasets that can be utilized for cross-domain DOD model training and evaluation. For each dataset, we provide the page images, bounding box annotations, PDF files, and the rendering layers extracted from the PDF files. Moreover, we propose a novel cross-domain DOD model which builds upon the standard detection model and addresses domain shifts by incorporating three novel alignment modules: Feature Pyramid Alignment (FPA) module, Region Alignment (RA) module and Rendering Layer alignment (RLA) module. Extensive experiments on the benchmark suite substantiate the efficacy of the three proposed modules and the proposed method significantly outperforms the baseline methods. The project page is at \url{https://github.com/kailigo/cddod}.
\end{abstract}

%%%%%%%%% BODY TEXT

\section{Introduction}
Document Object Detection (DOD) is the task of automatically decomposing a document page image into its structural and logical units (e.g., figures, tables, paragraphs). DOD is critical for a variety of document image analysis applications, such as document editing, document structure analysis and content understanding \cite{staar2018corpus,cattoni1998geometric,schreiber2017deepdesrt}.
Two popular document formats, image (e.g. scanned and camera-captured documents) and PDF, do not explicitly encode document structure:
images consist of pixels, while PDFs consist of vector, raster, and text marking operations that allow a document to be faithfully reproduced across devices (e.g., printers, and displays).

Though recent advances in object detection for natural scene images are impressive~\cite{ren2015faster,lin2017feature}, directly applying the same model to document images is likely sub-optimal due to large domain differences.
%Document images are characteristically different from natural scene images, so models shown to work well on natural scene images may not work well on document images. 
For example, document objects are more diverse in aspect ratio and scale than natural scene objects: tables may occupy a whole page, page numbers can be as small as a single digit, and a single line of text spanning the page has an extreme aspect ratio.
The intra-class variance of document objects is also usually larger than that of natural scene objects. 
Text can have arbitrary font face, style, position, orientation, and size.
Table cells can be filled with arbitrary content as long as they are aligned in a grid layout.
Document layouts and objects are also modular entities, so that, for example, examining the left half of a paragraph gives little information on how wide that paragraph is.
In contrast, missing parts of natural objects and scenes can be reasonably in-painted based on surrounding context~\cite{pathak2016context}.

% while generic object detection methods are  usually designed for nearly square objects. What’s more, the background spaces inside page objects maybe much larger than those between objects. This makes the grouping of object elements difficult \cite{li2018page,yi2017cnn}.  It  makes the texture based region proposal generation methods, such as selective search and region proposal network (RPN), not appropriate for page object detection task.

Another key challenge is that many factors influence the appearance of a document, such as document type (e.g. menu, scientific article), layout (e.g. Portrait vs Landscape or single/multi-column), and written language.
While learning a single model that can handle all varieties of documents is desirable, constructing such a comprehensive dataset appears infeasible.
Because labeled data is not available for every kind of document collection, we are motivated to examine cross-domain DOD.
In cross-domain DOD, we leverage labeled data in the source domain and unlabeled data in the target domain to train a detector for the target domain.
%It is infeasible to learn a model universally applicable for all kinds of document types: it is prohibited to get annotations for all kinds of documents and it is usually the case that we have no idea on the type of the coming document data. 
%This drives us to investigate cross-domain DOD, which has the goal of transferring knowledge from a labeled source domain to an unlabeled target domain such that a quality detector can be obtained for the target domain as well. 

To facilitate advancements in this field, we establish a benchmark suite on which cross-domain DOD models can be trained and evaluated. The benchmark suite consists of different types of document datasets; each serves as one domain. 
Each dataset is composed of the following components. (1) Document page images and bounding box annotations. These are essential data for detection model training and evaluation. (2) Raw PDF files used to generate the page images. The raw PDF files preserve the information lost when converting PDF pages to images, like the text and some meta-data. These extra sources of information may complement the visual information and benefit the detection task. (3) PDF rendering layers. A PDF page is in fact a mixture of text, vector, and raster content. These three content types can be rendered into separate layers for a PDF page, with each layer containing the pixel rendering arising only from one content type (text, vector, or raster). These rendering layers provide a structural abstraction of the PDF page and thus could be helpful for the detection task as well. 

In addition to the benchmark suite, we propose a novel cross-domain DOD model, which builds on top of the Feature Pyramid Networks (FPN) object detector \cite{lin2017feature} and addresses the domain shift problem by introducing three novel modules. The first module is the Feature Pyramid Alignment (FPA) module. FPA performs dense pixel-wise alignment between feature pyramids from source and target domains. Since each layer of the pyramids mixes both high- and low-level features, explicitly pushing cross-domain feature pyramid layers closer to each other achieves joint alignment of both low- and high-level semantics. The second module is the Region Alignment (RA) module. RA aims to enhance the alignment of semantically meaningful foreground regions from the two domains and explicitly pushes regions extracted from the two domains closer to each other. We adopt the focal loss \cite{lin2017focal} into our objective function to focus more on hard-to-align samples. The last module is the Rendering Layer Alignment (RLA) module. We utilize the PDF rendering layers available in our benchmark suite and generate for each page a mask which specifies the rendering layer that each pixel belongs to. RLA takes the mask as a kind of segmentation map for the page and trains an auxiliary segmentation task to further align the domains by predicting the masks from images of the two domains.

% Since the rendering layer masks are available and consistent for both the source and target domains, the auxiliary segmentation task shall benefit to further mitigate the domain gaps.

The contribution of this paper is three-fold: 
\begin{itemize}
	\vspace{-2pt}
	\item We establish a benchmark suite for cross-domain DOD model training and evaluation. To our best knowledge, we are the first to study this problem and the benchmark is the first one in this area. 
	\vspace{-4pt}
	\item We propose a novel cross-domain DOD model which introduces three novel modules to approach the domain shift problem. The three modules complement with each other and align the domains from both general image perspective and specific document image perspective. 
	\vspace{-4pt}
	\item Our model effectively mitigates the domain shift problem and significantly advances the baseline performance on the benchmark suite.
\end{itemize} 
% the baseline performance on Experiments shows that 
% builds up on the standard object detection neural networks and incorporates three novel modules to mitigate domain gaps. 
%We propose to utilize postage stamps associated with PDF pages to further bridge gaps between different document domains, by an auxiliary semantic segmentation task, which predicts postage stamps for images from both source and target domains.

\section{Related Work}
Our work is related to document object detection and cross-domain object detection for natural scene images.

\begin{figure*}
	\begin{center}
		\includegraphics[width=1.0\textwidth]{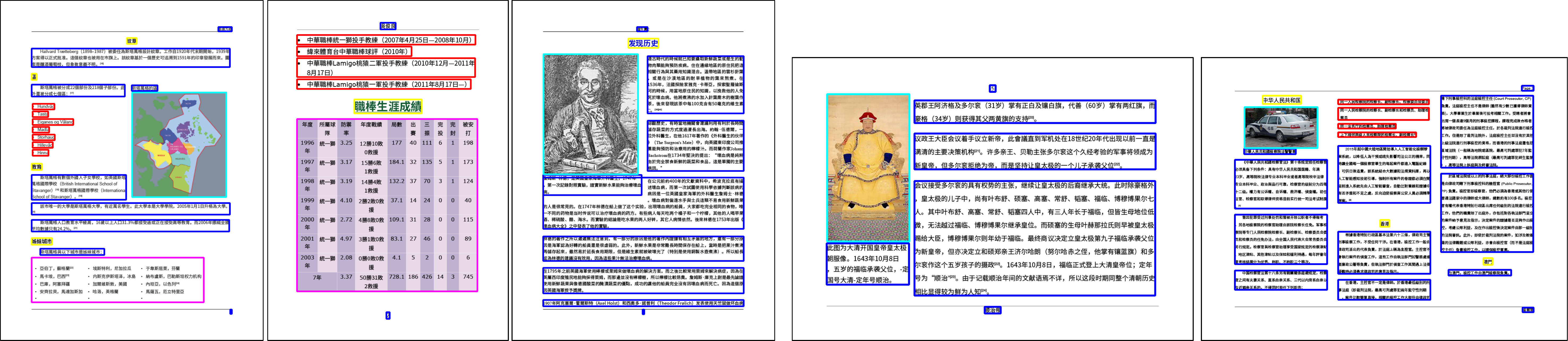}     
	\end{center}
	\vspace{-10pt}
	\caption{Samples from the \textit{Chn} dataset. Bounding box in colors are the ground truth labels: 
		\protect\cfbox{red}{list},
		\protect\cfbox{mymagenta}{table},
		\protect\cfbox{blue}{text},
		\protect\cfbox{mycyan}{figure}, and
		\protect\cfbox{yellow}{heading}.
	}
	\vspace{-10pt}
	\label{Chs_sample_combo}  
\end{figure*}

\subsection{Document Object Detection}
Most existing approaches to document object detection focus on certain types of objects, e.g., tables, figures, or mathematical formulas. Early works rely on various heuristic rules to extract and identify these objects from document images \cite{liu2017robust,garain2009identification,tran2015table}. These approaches often involve a set of hyper-parameters, which are difficult to adapt to new document domains. Recent works are usually data-driven and approach the problem with machine learning techniques, or a hybrid of heuristic rules and learning models. Taking advantage of the impressive progress of object detection on natural scene images, many works adapt natural image object detectors by considering the uniqueness of the document images  \cite{schreiber2017deepdesrt,gilani2017table}. 
He et al. \cite{he2017multi} propose a two-stage approach to detect tables and figures. In the first stage, the class label for each pixel is predicted using a multi-scale, multi-task fully convolutional neural network. Then in the second stage, heuristic rules are applied on the pixel-wise class predictions to get the object boxes. 
% class probabilities
% use a multi-scale, multi-task fully convolutional neural network (FCN) to predict region class probabilities at each pixel. 
% Then tables and figures are detected with some heuristic rules and a verification network.  
Gao et al. \cite{gao2017deep} utilize meta-data information of PDF files and detect formulas using a model that combines CNN and RNN.
% detect formulas from PDF files with the model that combine CNN and RNN
% combine CNN and RNN model to detect formulas from PDF files, but they require the meta-data information of PDF. 
A few works detect multiple types of document objects jointly in a single framework \cite{gao2017icdar2017}. 
Yi et al. \cite{yi2017cnn} adapt the region proposal approach and redesign the CNN architecture of common object detectors by considering the uniqueness of document objects. 
% redesign the region proposal method and network structure of traditional CNN based object detection methods to detect document objects.  
\cite{li2018page} first performs deep structure prediction and gets the
primitive region proposals from each column region.
Then, the primitive proposals are clustered and those within the same cluster are merged as a single object instance.
% and then cluster the primitive region proposals from each column region
% supervised clustering of the primitive region proposals from each column region, 
% and 

\subsection{Cross-Domain Object Detection}
Existing cross-domain object detection methods can be roughly divided into two categories: 
ones that are based on feature alignment and ones that are based on self-training. Methods in the former category train models from which domain-agnostic feature representations can be obtained for images from both domains \cite{zhu2019adapting,saito2019strong,chen2018domain,he2019multi}. To achieve this, these methods usually train domain classifiers and feature extraction models in an adversarial manner until the domain classifiers cannot distinguish the domains of the images from which 
the features are extracted.  The difference lies in the position and way that the domain classifiers are used. Methods in the latter category train the models recursively by generating pseudo bounding box labels for target images and updating the models with the generated pseudo labels \cite{kim2019self,inoue2018cross,khodabandeh2019robust,roychowdhury2019automatic}. Different methods vary in how they generate the pseudo labels or update the model. Both categories of methods benefit from style transfer techniques which first train a style transfer model (e.g., CycleGAN \cite{zhu2017unpaired}) using images from both domains and then apply the model to translate images from the source domain as the style of the target domain. Using this approach, labeled images of a similar style as that of the target domain can be obtained, which facilitates further domain alignment \cite{inoue2018cross,kim2019diversify}.

Our cross-domain DOD model inherits the merits of the recent cross-domain object detectors for natural scene images. We follow the line of approaches which addresses domain shifts by explicitly performing feature alignment. But instead of aligning either low-level features \cite{saito2019strong} or high-level features \cite{chen2018domain,zhu2019adapting}, or alignment them both separately \cite{he2019multi}, we align them jointly with the feature pyramids, as each layer of which is a mixture of both low and high-level features. Further, we propose a focal loss based region proposal alignment module which aims to enhance the alignment of foreground regions.  This module is different from the previous methods \cite{chen2018domain,he2019multi} that treat all region proposal equally, and instead focuses more on the hard-to-align proposals.  Moreover, we utilize the rendering layer data available for the document datasets and generate from them segmentation masks for both source and target images. We use the masks as additional cues to align the domains by training an auxiliary segmentation task.

\begin{figure*}
	\begin{center}
		\includegraphics[width=1.0\textwidth]{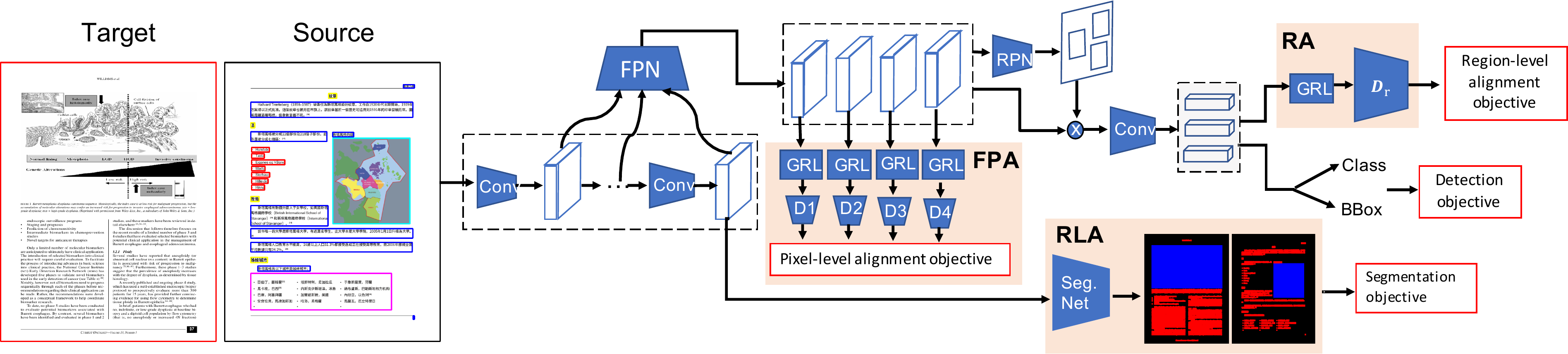}     
	\end{center}
	\vspace{-12pt}
	\caption{Summary of the proposed method. We build on FPN \cite{lin2017feature} and introduce three novel modules (masked in light pink) to align different document domains (English and Chinese in this example). The Feature Pyramid Alignment (FPA) module includes four binary domain classifiers $\{D_1, D_2, D_3, D_4\}$, corresponding to the four feature layers of the pyramid.  Each of these $D_j$ classifies pixels by image domain. The Region Alignment (RA) module is a binary domain classifier $D_i$, serving to classify the region proposals. The Rendering Layer Alignment (RLA) module is a segmentation network, which predicts rendering layer masks from the FPN layer. All binary domain classifiers follow a Gradient Reverse Layer (GRL)~\cite{ganin2015unsupervised}, which reverses the loss gradient during training and helps realize min-max optimization in each back-propagation.
	}
	\vspace{-10pt}
	\label{Framework}  
\end{figure*}

\section{Benchmark Suite}
There are a few existing datasets for document object detection. However, these datasets usually include annotations only for certain types of objects, e.g., table \cite{gobel2013icdar,fang2012dataset} or mathematical formulas \cite{lin2011mathematical}. \cite{gao2017icdar2017} established a dataset which contains annotations for three types of objects: table, figure and mathematical formula. However, this dataset is no longer publicly available. In addition, the largest existing dataset contains only 2000 page images \cite{gao2017icdar2017}, which is too small for modern deep object detectors.
% cannot meet the requirement of modern deep object detectors. 
% all these datasets are too small; 

Very recently, \cite{zhong2019publaynet} released a large-scale dataset for document object detection. It contains annotations for more than 3.5 million object instances from over 360 thousand page images extracted from medical journal articles. 
The annotated objects cover 5 classes: text, title, list, table and figure. The annotations are obtained automatically by matching the XML representations created by the publisher and the PDF content. We take advantage of this dataset and randomly select a subset, referred to as \textit{PubMed} in this work, for our cross-domain experiments. \textit{PubMed} includes 12871 images and 257830 bounding box annotations. We randomly split the dataset into 9653 images for training and 3218 images for testing.  
% Note how the "list" class is defined in the \textit{PubMed} dataset compared to the other datasets. In PubMed a "list" is a single region that contains all the numbered or bulleted items where as in the other datasets each element is considered its own region. for list instances in this dataset, the original annotation uses a bounding box covers all the items, which is not consistent with the case of the other datasets. 
% Note how the "list" class is defined in the \textit{PubMed} dataset
Note that the definition of the ``list'' class in \cite{zhong2019publaynet} is a single region that contains all the numbered or bulleted items. This definition is not consistent with that of the other datasets within the benchmark suite. We therefore preprocess the annotations of ``list'' in \cite{zhong2019publaynet} and split the ground truth bounding boxes into small ones for every individual items by detecting the list bullets or numbers.

%  in the \textit{PubMed} dataset is defined as  This is not consistent with the case of the other datasets. 

% \begin{figure}
% \begin{center}
%     \includegraphics[width=1.0\linewidth]{./figs/chs_sample_combo.pdf}     
% \end{center}
% \vspace{-10pt}
% \caption{Samples from the \textit{Chn} dataset. Bounding box in colors are the ground truth labels: 
% \protect\cfbox{red}{list-item},
% \protect\cfbox{mymagenta}{table},
% \protect\cfbox{blue}{text},
% \protect\cfbox{mycyan}{figure}, and
% \protect\cfbox{yellow}{heading}.
% }
% \label{Chs_sample_combo}  
% \end{figure}
% We To generate Simplified Chinese synthetic documents, we used 

% that given content it allows for natural looking well-tagged PDF generation. In our case, content is provided from 
Another dataset included in the benchmark suite is \textit{Chn}, a synthesized Chinese document dataset. It is generated by a tool which crawls Chinese Wikipedia pages and converts the content into natural looking well-tagged (bounding box annotations can be obtained from the tags) PDF files. Specifically, the tool converts each Wikipedia HTML page into a document by (a) randomly defining a layout to arrange the HTML content in document pages and (b) selecting the style of that content. The layout generation is controlled by a set of layout parameters that define the overall appearance and includes margins, number of columns, the white-space between the columns and the presence of headers / footers. Content is arranged based on the defined layout, which results in a Document Object Model (DOM) where most of the DOM elements correspond to tags generated in the final PDF. The styling parameters, which define the look and feel of paragraphs, headers etc., include font (family, size and style), as well as coloring schemes for lines (e.g., for tables). It should be noted that styling parameters follow a hierarchical pattern; for example defining the size of the base font automatically sets the font size for all the headers (h1, h2, etc.). Finally, to enforce the generated documents to be as natural looking as possible, the layout and styling parameters are randomly sampled from a distribution computed using real world document statistics.

After filtering out low quality samples, we obtain 8005 page images, with 203456 bounding box annotations for the same 5 classes as that of \textit{PubMed}. 
We further randomly select 5000 and 3005 page images for training and test, respectively. Figure \ref{Chs_sample_combo}  shows some samples from the dataset.

For the two datasets above, besides the images and the corresponding bounding box annotations, we also provide the raw PDF files used to generate the page images. Most meta-data are lost when converting PDF pages to images. We thus provide the source PDF files in our benchmark suite to enable future research to take advantage of these meta-data and advance the detection task or other relevant tasks, as Yang et. al. \cite{yang2017learning} did. They showed that the textual information in PDF files are helpful for document semantic structure extraction, when properly combined with the visual images. We also believe the textual information can also benefit the detection task for multi-modal (visual + textual) methods.

Moreover, we also provide the PDF rendering layers associated with the PDF pages. A PDF page is in fact rendered by a mixture of textual drawings, vector drawings, and raster drawings. Drawings of the same type lie in the same rendering layer, and we can extract the layers from PDF files. These rendering layers provide structural abstraction of PDF pages and thus shall be helpful for the detection task as well. 

We also utilize a human-annotated dataset for performance evaluation. The dataset includes 19355 page images and 257830 bounding box annotations for legal reports. We randomly select 9684 images for training and the remaining 9671 for testing.  This dataset is annotated with the same 5 classes as the other two datasets and we also utilize the rendering layers in this work.
We will refer this dataset as $\textit{legal}$ in future use.

\begin{figure*}
	\begin{center}
		\includegraphics[width=1.0\textwidth]{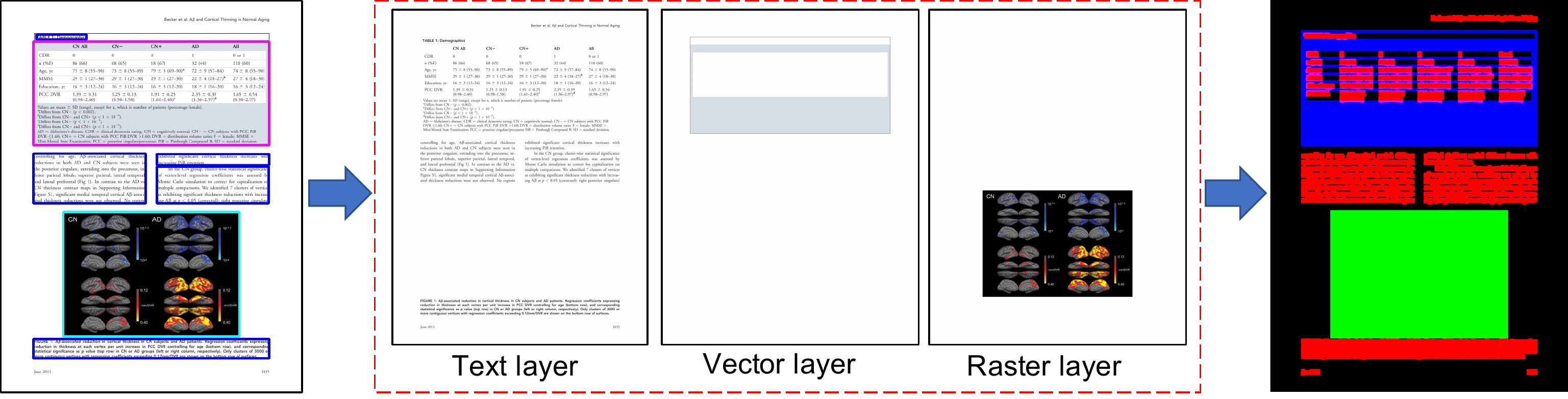}     
	\end{center}
	\vspace{-10pt}
	\caption{Generating the rendering layer mask from a PDF page.  Given a PDF page, we first use a tool to generate the text, raster and vector rendering layers, which are then binarized to separate the foreground from the background. Next, we merge the binary maps of the three layers and get a raw mask. Last, we perform morphological dilation and close operation to fill in the gaps between text characters and the holes in the raster drawings.}
	\label{layers}  
	\vspace{-10pt}
\end{figure*}

\section{Method}
Figure \ref{Framework} illustrates our proposed method. It is based on the Feature Pyramid Networks (FPN) and includes three novel domain alignment modules, namely, the Feature Pyramid Alignment (FPA) module, the Region Alignment (RA) module and the Rendering Layer Alignment (RLA) module. 

% During training, we are given a labeled dataset $\mathcal{S}=\{\mathcal{X}_s, \mathcal{Y}_s, \mathcal{P}_s\}$ from one domain and an unlabeled dataset $\mathcal{T}=\{\mathcal{X}_t, \mathcal{P}_t\}$ from another domain, where $\mathcal{X}_s$ and $\mathcal{X}_t$ are the images. $\mathcal{Y}_s$ are the bounding boxes of objects appearing in $\mathcal{X}_s$, specifying the locations and classes of the objects. $\mathcal{P}_s$ and $\mathcal{P}_t$ are the image postage stamps for the source and target datasets, respectively. Our proposed method manages to mitigate the domain gap between and $\mathcal{S}$ and $\mathcal{T}$ by the following Feature Pyramid Alignment (FPA), Region-level Alignment (RA) and Rendering Layer Alignment (RLA) techniques. Figure \ref{Framework} diagrams our framework% has semantics from low to high levels, and
% build a feature pyramid with high-level semantics throughout
% constructs feature pyramids in a  bottom-up pathway, a top-down pathway, and lateral connections
% Specifically, .
% Taking the standard ResNets \cite{he2016deep} as the backbone, 
% a bottom-up pathway results in 

\subsection{Feature Pyramid Networks}
FPN exploits the pyramidal feature hierarchy of convolutional neural networks and builds a feature pyramid of high-level semantics for all the layers. It is independent of the backbone convolutional architecture (we adopt the standard  ResNet-101 \cite{he2016deep} as the backbone). With the feature hierarchy $\{C_1, C_2, C_3, C_4\}$ from the layer1, layer2, layer3, and layer4 outputs of ResNet-101, FPN iterates from the coarsest feature map, up-samples it by a factor of 2 for the spatial resolution, and merges it (by element-wise addition) with the preceding map, which has undergone a 1$\times$1 convolution to reduce channel dimensions.
The merged feature map is then smoothed by a $3\times 3$ convolution to produce the final feature map. This iteration process outputs a feature pyramid $\{P_1, P_2, P_3, P_4\}$, where
\begin{equation}
P_i = conv3(up\_sample(P_{i+1}) + conv1(C_i)), \\\ i = 1, 2, 3, 4, 
\end{equation}
where $conv1$, $conv3$, and $up\_sample$ are 1$\times$1, 3$\times$3 and up-sampling operations, respectively. Note that $P_5$ is the result of 1$\times$1 convolution on $C_4$, i.e., $P_5=conv1(C_4)$.

Region proposals are extracted from all feature pyramid layers $\{P_1, P_2, P_3, P_4\}$ by the region proposal network (RPN). The obtained region proposals are then forwarded to the feature extraction module to obtain a feature vector for each proposal. For an image from the source dataset, we calculate the detection loss using  the bounding box ground truth:
\begin{equation}
L_{det}^s = L_{reg} (x^s, y^s) + L_{cls} (x^s, y^s),
\end{equation}
where $x^s$ and $y^s$ are the image and the ground truth annotation, respectively. The first term is the bounding box regression loss and the second term is the classification loss.

\subsection{Feature Pyramid Alignment}
% Our FPA module is inspired by the promotion of Feature Pyramid Network (FPN) \cite{lin2017feature} over the vanilla Faster RCNN \cite{ren2015faster} by constructing a
% feature pyramid where feature maps in all scales are semantically strong, including the high-resolution levels, which facilitates to detect objects in a wider range of sizes. Since 
As we can see above, feature maps in the pyramid are a mixture of both high- and low-level features; aligning feature pyramids from different domains therefore results in a joint alignment of both low- and high-level semantics. This is advantageous over existing methods where alignment is performed for low-level features only \cite{saito2019strong} or high-level features only \cite{chen2018domain}, or both of them separately \cite{he2019multi}.
Moreover, by virtue of building upon FPN, we inherit its strength of detecting objects of wide range of sizes, which is important for detecting objects in document images, as they can vary significantly in size. For example, ``text'' objects can occupy almost a whole page (e.g., long paragraphs), while others may be as small as a few characters or digits (e.g., page numbers or short section headings).

% Note this detection loss is only for source images, as we do not have  bounding box ground truth for target images. 

% the  and  can be calculated by using 
% Then, the region proposals are feed-forwarded normally to a 
% After obtained regions proposals from all the layers, they
% standard region proposal \cite{ren2015faster} can be performed on each layer of the pyramid. 
% standard object detection 
% On The result is a that has rich semantics at all levels

% \noindent\textbf{FPA}. 
% FPA takes advantages of the feature pyramid of FPN in incorporating low and high semantics in all scales and conducts feature alignment within the feature pyramid. In this way, we can jointly align both low and high semantics in all the scales, leading to better semantic alignment results. 
% 4 binary domain classifiers for feature maps in the pyramid. These domain classifiers 
Specifically, FPA includes 4 binary domain classifiers $\{D_1, D_2, D_3, D_4\}$ corresponding to $\{P_1, P_2, P_3, P_4\}$. These classifiers predict the domain labels (source or target) of the pixels in the feature maps.
We train the classifiers and FPN in an adversarial manner so that FPN is domain-agnostic once the domain classifiers cannot tell whether a pixel is from the source or target. 
To this end, we reverse the gradients with respect to $\{P_1, P_2, P_3, P_4\}$ to optimize the min-max problem in each individual back-propagation \cite{ganin2015unsupervised}. The loss function is as follows:
\begin{equation}
\begin{array}{cl}
\mathcal{L}_p = & -\frac{1}{4W^sH^s}\sum_{i=1}^{4}\sum_{w=1}^{W^s}\sum_{h=1}^{H^s}\log(D_i(P^s_{i, w, h})) - \\\vspace{4pt}
& \frac{1}{4W^tH^t}\sum_{i=1}^{4}\sum_{w=1}^{W^t}\sum_{h=1}^{H^t}\log(1-D_i(P^t_{i, w, h})),
\end{array}
\end{equation}
where $W^s$, $H^s$, $W^t$, and $H^t$ are the width and height of the source and target feature maps, respectively. $P^s_i$ and $P^t_i$ are the $i$-th layers of the source and target pyramids, respectively. 

% This strength is especially beneficial for detecting objects in documents, as they vary significantly in sizes. 

\subsection{Region Alignment}
The above FPA module performs pixel-wise dense alignment of the feature maps, which gives equal treatment to the foreground and background regions. However, we are more interested in the foreground regions, as they are more semantically meaningful to the detection task. Region proposals are the likely foreground regions, so we perform further alignment on them.

As shown in \cite{saito2019strong}, a ``weak global alignment'' of images 
from different domains results in better cross-domain detection performance, due to the focus on the ``hard-to-align'' images. The focal loss \cite{lin2017focal} is introduced to the domain classifier to give less weight to the easy-to-align images in the loss function. Inspired by this, we include focal loss in our region proposal domain classifier to focus more on the hard-to-align proposals. While \cite{saito2019strong} apply this strategy in image level, we do this at the region proposal level to emphasize the alignment of the foreground regions. 
It should be noted that while region proposal alignment has been investigated before in \cite{chen2018domain}, it treats all region proposals equally, which could cause the easy-to-align proposals to dominate the loss, leading to undesirable alignment results.
% Inspired by , which introduces focal loss \cite{lin2017focal} to the domain classifier to achieve ``weak global alignment'' of images from different domains, we instead apply focal loss on the classifier for region proposals 
% to have a ``weak alignment'' of the likely foreground regions. In this way, 
% the model is to focus more on the hard-to-align region proposals. 
% \cite{saito2019strong} has shown that weak alignment of the high-level semantics 
% With the same aim as \cite{saito2019strong}, 
% We differentiate from \cite{saito2019strong} that we apply  applies this in image level, while we do this on region proposal level.

With the introduction of the focal loss, our region alignment objective is as follows:
% It should be noted that any whole-image domain alignment technique can be applied on these region proposals. 
% Part of our solution is a new technique for domain-alignment specifically for object detection and is motivated by the fact that many of the object proposals are noisy or plain wrong, meaning that we can hurt the network performance by attempting to align the two domains on the noisy instances.  Using a DC weighted by the object-detection classification scores addresses this challenge by assigning low weights to proposals that the classifier is unable to classify with high confidence. This way, our objective function is as follows:
\begin{equation}
\begin{array}{cl}
\mathcal{L}_r = & -\frac{1}{R} \sum_{i=1}^{R} (1-D_r(r^s_i))^\gamma \log(D_r(r^s_i)) - \\
& \frac{1}{R} \sum_{i=1}^{R} (D_r(r^t_i))^\gamma \log(1-D_r(r^t_i)),
\end{array}
\end{equation}
where $R$ is the number of region proposal extracted; the terms $r^s_i$ and $r^t_i$ are the $i$-th region proposals extracted from the source and target images, respectively; $D_r$ is the binary domain classifier; and, $\gamma$ controls the weight on hard-to-align proposals.
% $w^s_i$ and $w^t_i$ are the weights for $r^s_i$ and $r^t_i$, respectively, 
% reflecting their possibility of being an ``object''. 
As in FPA, we reverse the gradients with respect to the proposals and execute adversarial training of the classifier and FPN in each individual back-propagation.

\subsection{Rendering Layer Alignment} 
The PDF pages are rendered into three separate layers, where each layer contains the pixels resulting from a single type of content: text, vector, or raster. These layers provide information about the content within a PDF page. More importantly, they are available and consistent for both source and target images. Thus, they can be used as an additional supervision cue to bridge domain gaps. RLA takes advantage of this and utilizes the rendering layers to generate for each page a mask which specifies the drawing type each pixel belongs to. Figure \ref{layers} illustrates this process.  

% To get the segmentation mask of a page, we first binarize the three rendering layers and merge the three binary maps. For pixels  
% we first binaries all 

%Since the rendering layer masks are available and consistent for both the source and target domains, the auxiliary segmentation task shall benefit to further mitigate the domain gaps.
%Thus, a postage stamp can be viewed as a kind of segmentation map of the corresponding page image. 

%RLA is to leverage this consistency as an additional supervision cues to bridge the gap between source and target. % and an auxiliary segmentation task is learned for image from both source and target domains. 

The mask can be viewed as a segmentation map for the page images and we can learn a model to predict the map from the image. Therefore, the
RLA module is a segmentation neural network which takes feature maps $C_4$  as input and outputs a dense possibility map of the drawing types of each pixel. The page masks are used as ground truth. Thus, the rendering layer segmentation objective is as follows:
% By minimizing the difference of the dense possibility map and the ground truth postage stamp map for both source and target images, the model is further regularized to be domain-agnostic.
\begin{equation}
\begin{array}{cl}
\mathcal{L}_s  = & -\frac{1}{W_m^sH_m^sC}\sum_{i=1}^{W_m^sH_m^s}\sum_{c=1}^{C}y_{i, c}\log p^s_{i, c} - \\
& \frac{1}{W_m^tH_m^tC}\sum_{i=1}^{W_m^tH_m^t}\sum_{c=1}^{C}y_{i, c}\log p^t_{i, c},
\end{array}
\end{equation}
where $W_m^s$, $H_m^s$, $W_m^t$, and $H_m^t$ are the width and height of the masks for the source and images, respectively; 
$p^s_{i,c}$ and $p^t_{i, c}$ are the probability of the $i$-th pixel of being class $c$; $y_{i, c}$ is the ground truth label; and, $C$ is the the number of classes. We find that the vector drawing class is not reliable, as vector drawings are usually too thin to have a concrete semantic meaning.  So, we merge it into the background class and keep ``background'', ``text'', and ``raster'' classes, i.e., $C=3$. 
% as each pixel can only categorized as text, vector or raster.

% There are often many layers producing output maps of the same size and we say these layers are in the same network \emph{stage}.
% For our feature pyramid, we define one pyramid level for each stage.
% We choose the output of the last layer of each stage as our reference set of feature maps, which we will enrich to create our pyramid.
% This choice is natural since the deepest layer of each stage should have the strongest features.

\subsection{Model Training and Inference}
The model is trained end-to-end by minimizing the sum of the above losses:
\begin{equation}
\mathcal{L} = \mathcal{L}_{det}^s + \lambda_1\mathcal{L}_{p} + \lambda_2\mathcal{L}_{r} + \lambda_3\mathcal{L}_{s},
\end{equation}
where $\lambda_1$, $\lambda_2$ and $\lambda_3$ are three hyper-parameters. 

For model inference, we remove the FPA, RA, and RLA modules and keep only the standard FPN. Then, the inference process is the same as the standard detection model: images are fed to the model and the detection bounding boxes are the output.

\begin{table}
	\small
	\renewcommand{\tabcolsep}{2.5pt}
	\begin{center}
		\begin{tabular}{|l|c|c|c|c|c|c|}\hline
			& text & list & heading & table & figure & MAP \\\hline
			FRCNN (source-only)
			& 61.7
			& 44.9
			& 75.2
			& 72.0
			& 65.4
			& 63.8
			\\
			% \hline
			SWDA \cite{saito2019strong}
			& 66.0
			& 23.3
			& 81.0
			& 85.1
			& 71.4
			& 65.3  
			\\\hline
			SWDA+RLA (Ours)
			& 67.4
			& 48.6
			& 82.9
			& 85.3
			& 59.3
			& \textbf{68.7}
			\\\hline
		\end{tabular}
	\end{center}
	\vspace{-5pt}
	\caption{Impact of adding the proposed RLA module on an existing work. The best results are in \textbf{bold}}
	\vspace{-5pt}
	% Effective of domain adaptation from \textit{Legal} to \textit{Chn} with postage stamps.
	\label{table_ablation_rl}
\end{table}

\section{Experiments}
% We conduct experiments on our proposed benchmark suite and compare with baseline methods. 
\noindent\textbf{Implementation details}. 
As introduced above, we build upon the standard feature pyramid networks (FPN) and propose three novel modules to combat domain shift problem. For FPN, we follow the most common practice and use ResNet-101 as the backbone. There are four domain classifiers in the proposed FPA module; they share the same structure but not the weights. We adopt the similar structure as \cite{saito2019strong} and use three convolution layers. The kernel size of the convolution layers is set as one and the padding size is zero. ReLU activation function is applied to the outputs of the first two convolution layers and Sigmoid is used for that of the last one. The RA module consists of three FC layers. ReLU and Dropout are applied on the outputs of the first two FC layers. 
% layer is appended to the first two The output of the local domain classifier is activated by Sigmoid function.  
For the segmentation network in the RLA module, we use the same structure as the DeepLab-V2 \cite{chen2017deeplab} and predict the segmentation mask from the feature map. 
% More details about the network architecture can be found in the supplementary material.

\begin{table}
	\small
	\renewcommand{\tabcolsep}{1.5pt}
	\begin{center}
		\begin{tabular}{|l|c|c|c|c|c|c|}\hline
			& text & list & heading & table & figure & MAP \\\hline
			FPN (source-only)
			& 60.9	
			& 51.5	
			& 74.6	
			& 69.6	
			& 67.8	
			& 64.9
			\\
			\hline
			% \multirow{3}{*}{Proposed} 
			FPN + FPA
			& 68.4
			& 51.9
			& 83.4
			& 68.1
			& 60.5
			& 66.5
			\\
			FPN + FPA + RA
			& 65.8
			& 52.5
			& 82.3
			& 74.8
			& 67.4
			& 68.6
			\\
			FPN + FPA + RA + RLA
			& 67.5
			& 53.6
			& 82.1
			& 76.6
			& 73.9
			& \textbf{70.7}
			\\\hline
		\end{tabular}
	\end{center}
	\vspace{-5pt}
	\caption{Ablation study about the effectiveness of the proposed components. The best results are in \textbf{bold}.
	}
	\vspace{-5pt}
	\label{table_ablation_final}
\end{table}

We train the networks with an SGD optimizer and an initial learning rate of 0.001, which is divided by 10 after every 8 epochs out of the total 12 epochs. For all our experiments, we set $\lambda_1=\lambda_2=0.1$ and $\lambda_3=0.01$. 
Following \cite{saito2019strong}, the focal loss parameter is set as $\gamma=5.0$.
% for 50K iterations, then with learning rate 0.0001 for 20K more iterations and reported the final performance. All models are trained with this scheduling and we reported the performance trained after 70K iterations. Wi
% thout specific notation, we set λ as 1.0 and γ as 5.0. 

In all cross-domain experiments, we use the training splits of the source and target datasets for training and evaluate on the test split of the target dataset. During training, only the labels for the source dataset are available. We set the shorter side of the image to 600 pixels. and report mean average precision (MAP) with a threshold of 0.5 to evaluate different methods.
We implemented all methods with PyTorch \cite{paszke2017automatic}.

% The learning rate is 0.02 for 12 epochs
% ,

\begin{table*}
	\small
	\renewcommand{\tabcolsep}{4pt}
	\begin{center}
		\begin{tabular}{|l|cccccc|cccccc|} \hline		
			& \multicolumn{6}{|c|}{\textit{Legal} $\rightarrow$ \textit{Chn}}   & \multicolumn{6}{|c|}{\textit{Chn} $\rightarrow$ \textit{Legal}} \\\hline
			& text 
			& list 
			& heading 
			& table 
			& figure 
			& MAP  
			%%%%
			& text 
			& list 
			& heading 
			& table 
			& figure 
			& MAP 
			\\\hline
			Oracle
			& 90.5
			& 89.9
			& 90.5
			& 88.9
			& 90.5
			& 90.1
			& 84.5
			& 88.8
			& 82.4
			& 78.6
			& 71.9
			& 81.3
			\\\hline
			FRCNN (source-only)
			& 73.7
			& 57.9
			& 74.8
			& 66.2
			& 76.5 
			& 69.8
			& 60.7
			& 50.9
			& 30.7
			& 47.2
			& 24.1
			& 42.7
			\\
			% \hline
			FPN (source-only)
			& 75.0
			& 67.3
			& 80.3 
			& 65.1
			& 85.2
			& 74.6
			& 59.0
			& 54.5
			& 26.4
			& 53.2
			& 24.7
			& 43.6
			\\
			% \hline
			SWDA \cite{saito2019strong}
			& 74.9
			& 67.7
			& 73.8
			& 74.0
			& 86.6
			& 75.4  
			& 52.2
			& 51.1
			& 31.9
			& 58.1
			& 29.9
			& 44.6
			\\\hline
			SWDA + RLA (Proposed)
			& 75.4
			& 73.2
			& 79.1
			& 78.7
			& 87.7
			& \textbf{78.8}
			& 59.2
			& 57.0
			& 33.0
			& 56.0
			& 28.9
			& 46.8
			\\ 
			% \cline{2-14}
			Proposed
			& 76.8
			& 75.5
			& 79.2
			& 72.5
			& 88.2
			& 78.5
			& 62.7
			& 62.3
			& 35.5
			& 57.9
			& 26.9
			& \textbf{49.1}
			\\\hline
		\end{tabular}
	\end{center}
	\vspace{-5pt}
	\caption{Cross-domain detection results between \textit{Legal} to \textit{Chn}. The ``Oracle'' results are obtained by FPN trained with labeled training data of the target domain. The best results are in \textbf{bold}.
	}
	\label{table_comparison_legal_chs}
	\vspace{-5pt}
\end{table*}

\begin{table*}
	\small
	\renewcommand{\tabcolsep}{4pt}
	\begin{center}
		\begin{tabular}{|l|cccccc|cccccc|} \hline		
			& \multicolumn{6}{|c|}{\textit{Chn} $\rightarrow$ \textit{PubMed}}   & \multicolumn{6}{|c|}{\textit{PubMed} $\rightarrow$ \textit{Chn}} \\\hline
			& text 
			& list 
			& heading 
			& table 
			& figure 
			& MAP  
			%%%%
			& text 
			& list 
			& heading 
			& table 
			& figure 
			& MAP 
			\\\hline
			Oracle
			& 90.6	
			& 68.3	
			& 90.3	
			& 90.7	
			& 90.7	
			& 86.1	
			& 90.5
			& 89.9
			& 90.5
			& 88.9
			& 90.5
			& 90.1
			\\\hline
			FRCNN (source-only)
			& 41.3
			& 14.3
			& 45.4
			& 67.4
			& 57.4
			& 45.2
			& 26.6
			& 17.7
			& 19.6
			& 45.5
			& 51.9
			& 32.3
			\\
			% \hline
			FPN (source-only)
			& 47.2
			& 19.5
			& 47.1
			& 64.3
			& 64.7
			& 48.6
			& 38.4
			& 25.0
			& 26.7
			& 45.9
			& 28.7
			& 32.9
			\\
			% \hline
			SWDA \cite{saito2019strong}
			& 56.0
			& 20.3
			& 52.2
			& 81.2
			& 44.5
			& 50.9
			& 53.0
			& 18.5
			& 35.0
			& 64.7
			& 64.3
			& 47.1
			\\\hline
			SWDA + RLA (Proposed)
			& 50.6
			& 24.3
			& 50.5
			& 74.6
			& 59.2
			& 51.8
			& 48.9
			& 25.3
			& 39.8
			& 60.0
			& 74.3
			& 49.7
			\\
			% \cline{2-14}
			Proposed
			& 55.8
			& 28.6
			& 54.1
			& 79.6
			& 52.5
			& \textbf{54.1}
			& 36.7
			& 44.4
			& 42.1
			& 64.3
			& 79.4
			& \textbf{53.4}
			\\\hline
		\end{tabular}
	\end{center}
	\vspace{-5pt}
	\caption{Cross-domain detection results between \textit{PubMed} to \textit{Chn}.}
	\label{table_comparison_pubmed_chs}
	\vspace{-5pt}
\end{table*}

\begin{table*}[!htbp]
	\small
	\renewcommand{\tabcolsep}{4pt}
	\begin{center}
		\begin{tabular}{|l|cccccc|cccccc|} \hline		
			& \multicolumn{6}{|c|}{\textit{Legal} $\rightarrow$ \textit{PubMed}}   & \multicolumn{6}{|c|}{\textit{PubMed} $\rightarrow$ \textit{Legal}} \\\hline
			& text 
			& list 
			& heading 
			& table 
			& figure 
			& MAP  
			%%%%
			& text 
			& list 
			& heading 
			& table 
			& figure 
			& MAP 
			\\\hline
			Oracle
			& 90.6
			& 68.3
			& 90.3
			& 90.7
			& 90.7
			& 86.1
			& 84.5
			& 88.8
			& 82.4
			& 78.6
			& 71.9
			& 81.5
			\\\hline
			FRCNN (source-only)
			& 61.7
			& 44.9
			& 75.2
			& 72.0
			& 65.4
			& 63.8
			& 37.3
			& 37.3
			& 27.1
			& 29.8
			& 8.3
			& 28.0
			\\
			% \hline
			FPN (source-only)
			& 60.9	
			& 51.5	
			& 74.6	
			& 69.6	
			& 67.8	
			& 64.9	
			& 35.3
			& 41.4
			& 28.5
			& 30.5
			& 3.7
			& 27.8
			\\
			SWDA \cite{saito2019strong}
			& 66.0
			& 23.3
			& 81.0
			& 85.1
			& 71.4
			& 65.3
			& 37.3
			& 36.1
			& 44.0
			& 48.5
			& 10.5
			& 35.3
			\\
			\hline
			% \multirow{2}{*}{Ours}  
			SWDA + RLA (Proposed)
			& 67.4
			& 48.6
			& 82.9
			& 85.3
			& 59.3
			& 68.7
			& 36.8
			& 39.0
			& 43.4
			& 50.7
			& 11.9
			& \textbf{36.4}
			\\
			% \cline{2-14}
			Proposed
			& 67.5
			& 53.6
			& 82.1
			& 76.6
			& 73.9
			& \textbf{70.7}
			& 37.1
			& 49.6
			& 42.5
			& 31.1
			& 12.0
			& 34.5
			\\\hline			
		\end{tabular}
	\end{center}
	\vspace{-5pt}
	\caption{Cross-domain detection results between \textit{Legal} and \textit{PubMed}.}
	\label{table_comparison_pubmed_legal}
	\vspace{-5pt}
\end{table*}

\subsection{Ablation Study}
To address the domain shift problem, we propose three novel modules on top of the standard object detection model, namely, feature pyramid alignment (FPA) module, Region Alignment (RA) module, and the Rendering Layer Alignment (RLA) module. To evaluate the effectiveness and impact of the three modules, we conduct an ablation study on the adaption from \textit{Legal} and \textit{PubMed}. 

\noindent\textbf{RLA}. RLA takes the rendering layers available in both source and target domains as additional alignment cues and trains the network with an auxiliary segmentation task.  To evaluate its effectiveness, we first attach it to a recent cross-domain object detection model SWDA \cite{saito2019strong} and evaluate the resulting performance. The results in Table \ref{table_ablation_rl} show that the MAP is boosted by 3.2 points with this module added. Furthermore, as Table \ref{table_ablation_final} shows, when used as a component of our proposed method, RLA raises the MAP from 68.6 to 70.7. These consistent performance gains substantiate RLA's effectiveness. 

\noindent\textbf{FPA}. FPA performs domain alignment by pushing the feature pyramids of images from different domains closer together. Since each layer of the feature pyramids incorporates both low and high features, FPA thus jointly aligns low- and high-level semantics. 
% Besides, extracting region proposals on the feature pyramids is also beneficial for detecting objects of wide range of sizes. 
Table \ref{table_ablation_final} shows that 
% by extracting region proposals from feature pyramids, the detection capability of FRCNN on the target dataset is been significantly boosted when trained with only labeled data from the source domain, with MAP raised from 69.8 to 74.6.  This is reasonable as document objects vary significantly in size. Moreover, 
FPA leads to a gain of 1.6 MAP relative to the FPN baseline. 
% the performance is further boosted by nearly 3 points, thus substantiating its effectiveness. 

\noindent\textbf{RA}. RA enhances the alignment of foreground regions by aligning the extracted region proposals, with a focal loss based learning objective to focus more on the hard-to-align ones. Table \ref{table_ablation_final} shows that it raises MAP from 66.5 to 68.6.

\begin{figure*}
	\begin{center}
		\includegraphics[width=1.0\textwidth]{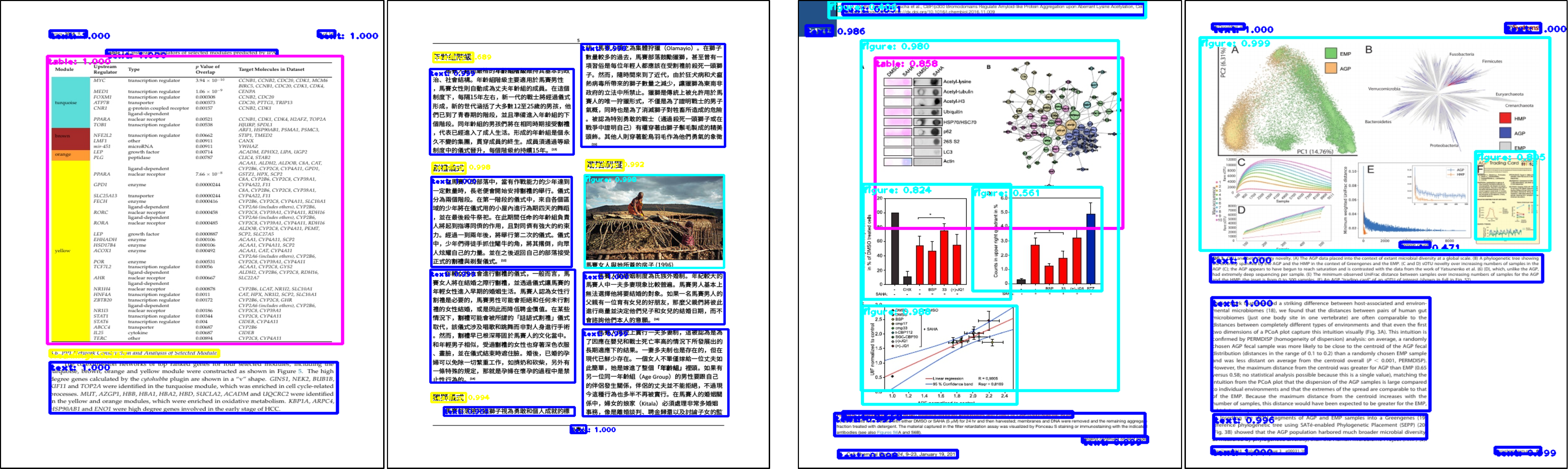}     
	\end{center}
		\vspace{-10pt}
	\caption{Sample detection results. }
	\label{result_samples}  
	% 		\vspace{-10pt}
\end{figure*}

\subsection{Comparative Results}
We conduct cross-domain evaluation between the three datasets, \textit{Chn}, \textit{Legal} and \textit{PubMed}. The first one is a Chinese document dataset and the latter two are English datasets. We first conduct cross-lingual performance evaluation between \textit{Chn} and \textit{Legal}, and between \textit{Chn} and \textit{PubMed}. Table \ref{table_comparison_legal_chs} and Table \ref{table_comparison_pubmed_chs} shows the experimental results.
Since \textit{Legal} and \textit{PubMed} belong to different English document categories, there is a domain gap between them. Therefore, we also conduct cross-category detection evaluations between these two datasets. Table \ref{table_comparison_pubmed_legal} shows the results.

We observe similar behavior across the three tables. The FPN baseline usually outperforms the FRCNN baseline. This is because document objects vary significantly in size, which FPN is more able to deal with. SWDA builds on top of FRCNN, leading to consistent performance gains. This shows that some domain adaptation techniques are applicable for various image types. Adding the document-specific alignment module RLA to SWDA results in consistent performance gains for all cases. This substantiates the ability of RLA to mitigate the domain shift problem. Our proposed method builds upon FPN and introduces three novel components. The tables demonstrate that it significantly outperforms the FPN baseline and also SWDA for almost all the cases. In addition, with the RLA module, our method surpasses SWDA for almost all experiments. This suggests that our proposed FPA and RA modules are superior to their counterparts in SWDA. Despite these improvements, the results are still far lower than the oracle setting in which FPN is trained on labeled data from the target domains. This shows that domain shifts are indeed a serious problem for DOD and they have not been fully addressed.  

One may have noticed from Table \ref{table_comparison_pubmed_chs}
that there is a huge performance jump for SWDA over the FRCNN baseline when adapting \textit{PubMed} to \textit{Chn}: MAP is improved from 32.3 to 47.1. 
Similarly, in Table \ref{table_comparison_pubmed_legal}, SWDA raises MAP over the FRCNN baseline from 28.0 to 35.3 when adapting \textit{PubMed} to \textit{Legal}.
We believe the reason for such huge gains is that \textit{PubMed} is comprised of page images of scientific journal articles that share similar formatting templates. So, the diversity of this dataset is limited. A model trained with labeled data of this dataset should be able to effectively handle other data within the dataset, as evidenced by the high oracle results, but it is less likely to generalize to other datasets. This problem is not as severe for the other two datasets, \textit{Chn} and \textit{Legal}, because the diversity is carefully considered when synthesizing \textit{Chn} and selecting data of \textit{Legal} for annotation. So, the impact of domain adaption is much more significant for \textit{PubMed} than the other two datasets when they are used as the source domain.

\begin{table}
	\small
	\renewcommand{\tabcolsep}{1.5pt}
	\begin{center}
		\begin{tabular}{|l|c|c|}\hline
			& \textit{Kitti} $\rightarrow$ \textit{Cityscape} & \textit{Cityscape} $\rightarrow$ \textit{Kitti}  \\\hline
			SWDA \cite{saito2019strong} 	&  41.8 & 70.6 \\\hline
			Proposed &42.9 & 73.3 \\\hline
		\end{tabular}
	\end{center}
	\vspace{-5pt}
	\caption{Cross-domain detection results for natural scene images.}
	\label{exp_natural_images}
	\vspace{-8pt}
\end{table}

\subsection{Further Analysis}
\noindent\textbf{Experiments on natural images}.
Without the proposed RLA module, which is specially designed for the DOD task, our method (FPN plus the FPA and RA modules) can be applied on natural scene images as well. 
Following the previous methods \cite{chen2018domain,he2019multi}, we conduct cross-domain ``car'' detection evaluation on the  \textit{Cityscape} \cite{cordts2016cityscapes} and \textit{Kitti} \cite{geiger2013vision} datasets. There are 14999 images in the \textit{Kitti} dataset and we chose 7481 images in the training set for both adaptation and evaluation. The \textit{Cityscape} dataset has 3475 images and we use 2975 images for adaptation training and the remaining 500 images for evaluation.
% in the training set, and 500 images in the validation set. We use the unlabeled images from the training set as the target domain to adapt our detector, and the results are reported on the validation set. 
The results in Table \ref{exp_natural_images} show that our method also outperforms SDWA for the natural scene image cross-domain detection task, especially for adaptation from \textit{Cityscape} and \textit{Kitti}, where we achieve 2.7\% improvement for the AP of car. This set of experiment further substantiates the efficacy of the proposed adaptation modules.

\noindent\textbf{Visualization of detection results}.
Figure \ref{result_samples} visualizes some detection results from \textit{Chn} and \textit{PubMed}. We can see that the proposed method in most cases can successfully decompose a complex page into semantically meaningful regions, with high localization precision and confident classification scores for objects of extremely diverse sizes. For example, in the first image, both the large table which covers about two-third of the page and the tiny pagination are perfectly detected.  However, the proposed method tends to make mistakes for ambiguous objects whose semantic meanings can be correctly determined only within the context. For example, in the fourth (right-most) image, there is a composite figure comprised of six sub-figures. Each sub-figure alone is a figure instance. But when considering the context, it is wrong to detect them as object instances individually. Similar cases also appear in the third image.

\section{Conclusion}
We investigate cross-domain document object detection by proposing a benchmark suite and a novel method. The benchmark suite includes different types of datasets on which cross-domain document object detectors can be trained and evaluated. For each dataset, we provide not only the essential components, page images and bounding boxes annotations, but also auxiliary components, raw PDF files and the PDF rendering layers. 
% Those auxiliary components could be helpful for the detection task. 
% The essential components are necessary for cross-domain document object training and test, while the auxiliary components could help boost the detection performance. 
The proposed model builds upon the standard object detection model with three novel domain alignment modules, namely, the feature pyramid alignment (FPA) module, the Region Alignment (RA) module, and Rendering Layer Alignment (RLA) module. 
% Experimental reuslts 
% FPA performs joint low and high-level semantics alignment by explicitly aligning the feature pyramids of images from different domains. RA enhances the alignment of the semantically meaningful foreground regions by alignment the region proposals extracted from feature pyramids. RLA generates a page segmentation mask from the rendering layers for both source and target images, and further align the domains by training an auxiliary segmentation task by predicting the segmentation mask from feature maps. 
Experiments on the benchmark suite confirm the effectiveness of the proposed novel components and that the proposed method significantly outperforms the baseline methods. In addition, the proposed method also improves over the state-of-the-art method for cross-domain object detection on natural scene images. 

\vspace{5pt}
\noindent\textbf{Acknowledgement}: This work was partially done during the internship of the first author at Adobe Research and was partially supported by Adobe Research funding. We thank Richard Cohn and Kana Sethu for coding the tool and instructing how to use it for synthesizing documents.

%Despite of the significant improvements over the baseline, our results are far lower than the oracle ones. This means the domain shift problem has not been well addressed, which calls for future investigation. 

\clearpage
%\breakpage
{\small
	\bibliographystyle{ieee_fullname}
	\bibliography{egbib}
}

	\end{document}